\documentclass[preprint,review,5p]{elsarticle}

\usepackage{amssymb}

\usepackage{amsmath}
\usepackage{graphicx}
\usepackage{natbib}
\usepackage{tabularx}
\usepackage{framed}
\usepackage{lineno}
\usepackage{enumerate}
\usepackage{lmodern}
\usepackage{float}

\journal{Machine Learning with Applications}
\begin{document}

\begin{frontmatter}



\title{Incremental hierarchical text clustering methods: a review}


\author[ufla]{Fernando Simeone}\ead{fsimeone@posgrad.ufla.br}
\author[ufla]{Maik Olher Chaves}\ead{maikchaves@posgrad.ufla.br}
\author[ufla]{Ahmed Esmin}\ead{ahmed@ufla.br}

\address[ufla]{Department of Computer Science, Federal University of Lavras (UFLA), Campus Universitario, O Box 3037, Lavras MG 37200-000, Brazil}


\begin{abstract}
The growth in Internet usage has contributed to a large volume of continuously available data, and has created the need for automatic and efficient organization of the data. In this context, text clustering techniques are significant because they aim to organize documents according to their characteristics. More specifically, hierarchical and incremental clustering techniques can organize dynamic data in a hierarchical form, thus guaranteeing that this organization is updated and its exploration is facilitated. Based on the relevance and contemporary nature of the field, this study aims to analyze various hierarchical and incremental clustering techniques; the main contribution of this research is the organization and comparison of the techniques used by studies published between 2010 and 2018 that aimed to texts documents clustering. We describe the principal concepts related to the challenge and the different characteristics of these published works in order to provide a better understanding of the research in this field.

\end{abstract}

\begin{keyword}
Data mining \sep hierarchical clustering \sep incremental clustering \sep text clustering \sep text mining

\end{keyword}

\end{frontmatter}



\section{Introduction}
\label{introducao}

Internet usage has grown significantly in the last decade, thus leading to an increase in the amount of available data and the frequency at which these data are updated. This situation can be easily observed on news websites and in social media. The increasing amount of available information has resulted in the necessity to organize this information such that it can be easily found, explored, and analyzed. In this sense, one of the main ways that humans use to handle the large amounts of data is to classify them or group them into categories according to their characteristics \citep{Jain:1999}.

Text clustering is challenging because of the computational complexity that should be linear with respect to the number of dimensions (terms) and the number of clusters that are unknown at the beginning, so the algorithm has to be able to identify them through several topics \citep{Garcia:2010a}. Due the large number of text documents, the algorithms used must be efficient and scalable. Hierarchical clustering technique is highly used on text clustering algorithms due the intuitive way to organize and analyze the topics and context of text data \citep{cui:14}. 

In order to address that challenges, researchers have focused on the study of automated techniques to group data that have similar characteristics; these techniques are known as clustering techniques. Unlike supervised classification, clustering techniques do not use a training dataset, and the categorization of the data is obtained by analyzing the characteristics of that data itself. In clustering algorithms, the data are grouped according to the degree of similarity between them \citep{Cherkassky:2007}. Such techniques have proven to be suitable in scenarios that lack prior knowledge of the data or those that pose challenges in the creation of a suitable training dataset.

In order to improve data organization, it is appropriate to use hierarchical clustering algorithms, which organize the data into different levels of abstraction. In this type of algorithm, the resulting clusters are organized as a tree to facilitate the exploration and visualization of data and clusters \citep{Zhao:2005}.

The dynamic nature of the Web makes this problem even more challenging. In contexts such as news websites, information is constantly added and archived. In such a situation, traditional clustering algorithms are not effective because, for execution, they require the entire database to be available. However, incremental clustering algorithms can deal with dynamic situations. They can process the addition or removal of data by adapting the existing grouping structures as necessary \citep{Garcia:2010a}.

Owing to the relevance of this topic, a search was made on the main digital libraries (ACM, Science Direct and IEEE Xplore) and there was not found surveys comparing recent clustering techniques for text documents. To the best of our knowledge, this is the first surveys about incremental hierarchical clustering methods for text. Therefore, the aim of this study is to gather and compare recent studies that have developed or used incremental hierarchical clustering techniques for text documents. 

The rest of this paper is organized as follow: In Section \ref{sec:tecnicas_de_agrupamento}, the main concepts of clustering techniques are described. 
In Section \ref{sec:pesquisa_sistematizada}, the methodology used to obtain recently published works is presented. Section \ref{sec:trabalhos_encontrados} offers a survey about the previous work done on the hierarchical incremental text clustering algorithms and applications. 
The sections \ref{sec:trabalhos_encontrados_aplicacoes} and \ref{sec:trabalhos_encontrados_tecnicas} describe these studies. The similarity and distance measures used in these studies are listed in Section \ref{sec:medidas_similaridade}, the most commonly used databases in the considered studies are enumerated in Section \ref{sec:bases_de_dados}, the metrics used to evaluate the generated clusters are described in Section \ref{sec:metricas}, the baseline algorithms used in the studies are mentioned in Section \ref{sec:algoritmos_comparados} and a discussion about the characteristics of the studied papers is presented in Section \ref{sec:discussao}. 
In Section \ref{sec:conclusoes}, the final considerations and the conclusions of this study are stated.


\section{Clustering Techniques}
\label{sec:tecnicas_de_agrupamento}

The clustering is an unsupervised learning approach without labeled data. The clustering problem consists in the separation of data into groups, also known as clusters, based on a similarity measure \citep{liu:2013}. 
Such tasks could also be defined as an unsupervised learning of patterns, such as observations or data samples, in clusters. 
These data are generally represented as vectors or points in multidimensional space, and are grouped according to their similarities \citep{Jain:1999}. 
The resulting groups (clusters) consist of items that are similar to each other and dissimilar to items placed in other clusters \citep{Berkhin:2006}.
Documents on clustering algorithms usually belong to only one cluster, but some techniques allow intersection between clusters, so the documents may belong to more than one group. These techniques form what are known as overlapping clustering \citep{saxena:17}. 

As explained by \citet{Feldman:2006}, there are differences between supervised classification tasks and clustering tasks. In a supervised classification task, a set of classified and labeled data is already known. This dataset is used in the algorithm training process to enable the algorithm to learn the patterns in the data; thus, the algorithm is able to classify new data. However, in a clustering task, clusters with some meaning are generated from a set of unlabeled data, with no prior knowledge.

According to \citet{Jain:2010}, clustering techniques have been used with three main objectives: (1) identifying implicit information, such as anomalies and salient characteristics in the data; (2) natural classification, to identify the degree of similarity between data; and (3) facilitating understanding in the form of data organization and summarization through prototypes.

Prototypes are representations of the clusters that describe their elements in a compact manner; they facilitate human understanding about the grouping and optimize the subsequent processing of the data. Such information is important because clustering techniques are applied to datasets that are not previously labeled \citep{Jain:1999}. When the data are represented as points in space, the prototype can be represented as a central point in the cluster. In the case of textual documents clustering, the prototype could contain a list of the most frequent words from the cluster \citep{saxena:17}.

In order to build the clusters, elements that end up in the same group are more similar to each other than to the elements in other groups. The elements must be as homogeneous as possible related to its group and as distinct as possible related to elements allocated to other groups \citep{cornuejols:18}. There are different methods that can be used determine the degree of similarity or dissimilarity between the data. As explained by \citet{Cherkassky:2007}, such measures are known as similarity measures; they are selected subjectively in order to obtain interesting clusters for the given data context. It may be difficult to define a method to measure the similarity between two items in a dataset to be analyzed. According to \citet{Huang:2008}, the definition of similarity or dissimilarity between documents is not always clear, and it normally varies according to the context of the actual problem. Section \ref{sec:medidas_similaridade} gives more details about some of the most commonly used similarity and distance measures.

\citet{Xu:2009} and \citet{Jain:1999} explain that clustering techniques are generally divided into two categories: partitional clustering and hierarchical clustering. Partitional clustering algorithms organize the data into a pre-established number of groups without a hierarchical structure by identifying the partitions that optimize the clustering criteria. As defined by \citet{Hansen:1997}, given a sample ${D=\{d\textsubscript{1},d\textsubscript{2},...,d\textsubscript{N}\}}$ of ${N}$ elements to be clustered, the partitional cluster seeks to establish a set of $\mathit{M}$ partitions ${P_m=\{C_1,C_2,...,C_M\}}$, such that:

\begin{enumerate}[(a)]

  \item  \quad  ${C_j \neq \emptyset  \quad   j=1,2,...,M}$

  \item  \quad  ${C_i \cap C_j = \emptyset  \quad   i,j=1,2,...,M \quad i \neq j}$
  
  \item  \quad  ${\bigcup_{j=1}^{M}{C_j} = D}$
\end{enumerate}

The generated clusters cannot be empty and cannot have intersections between them. However, hierarchical clustering algorithms generate nested clusters, such as sets and subsets, a result that is achieved based on the criteria to join or divide each cluster. Hierarchical clustering is described in the next section.

\subsection{Hierarchical Clustering}
\label{sec:tecnicas_de_agrupamento_hierarquico}

Hierarchical clustering techniques are characterized by generating clusters of data with nested partitions (subgroups) \citep{Xu:2009}. This cluster hierarchy can be represented by a tree, which provides a view of the data at different levels of abstraction. This structure containing data clusters with different levels of granularity is ideal for the interactive exploration and visualization of the data \citep{Zhao:2005}. 

Hierarchical clustering are classified into agglomerative and divisive. Both categories are different ways to find the most efficient step at each stage. This technique is suitable for clustering a large volume of data and present the data in a tree structure organized from data similarities. One characteristic of this technique is that it does not need a prior information about the number of clusters, this characteristic is useful on text clustering.  

\begin{figure}[ht!]
  \centering
  \includegraphics[width=9cm]{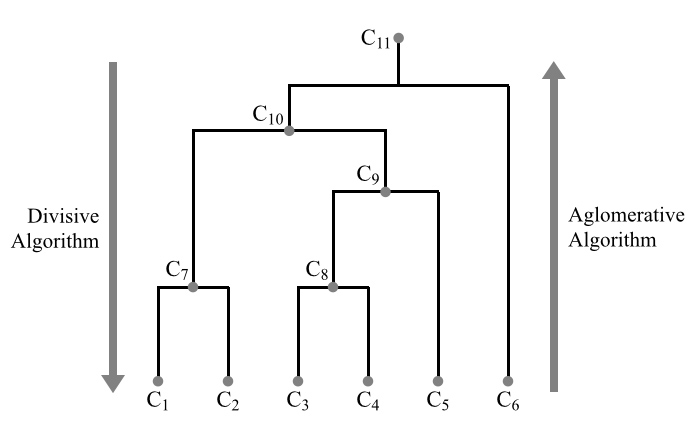}
  \caption{Example of dendrogram
  graph generated from hierarchical clustering, adapted from \citet{Xu:2009}}
  \label{fig_dendrograma_agrupamento}
\end{figure}

The two categories of hierarchical clustering differ into the way they proceed. Agglomerative algorithms begin by creating one cluster per document, and, during the iterations of the algorithm, these clusters are joined to eventually form a single cluster, which contains all the subgroups and documents. 
Divisive algorithms work in the opposite manner, beginning with a single cluster containing all the documents, and, during the iterations, the clusters are divided into subgroups until only one element remains per cluster \citep{Feldman:2006}. 
Figure \ref{fig_dendrograma_agrupamento} shows an example of a hierarchy generated by clustering techniques. In this example, the agglomerative algorithms generate a bottom-up hierarchy in the dendrogram graph, while the divisive algorithms use the reverse approach.

Agglomerative algorithms are more widely used for hierarchical clustering. The general behavior of these algorithms can be described by the following steps \citep{Xu:2005}:

\begin{enumerate}
  \item Start with $N$ clusters, each containing one document. Calculate the distance matrix for the $N$ clusters, containing the distance of each cluster from all the others.

  \item Given the distance function $\mathit{Dist(*,*)}$, perform a search in the distance matrix for the minimum value:

  \begin{equation}
    \nonumber
    Dist(C_{i},C_{j}) = \min_{\substack{1 \leq m, l \leq N \\ m \neq 1 }}  Dist(C_{m},C_{l})
  \end{equation}  

  Then, combine $\mathit{C_{i}}$ and $\mathit{C_{j}}$ to form a new cluster.

  \item Update the distance matrix to include the newly created cluster and calculate its distances from all the other existing groups.

  \item Repeat steps 2 and 3 until all the documents are in the same cluster.
    
\end{enumerate}

Divisive algorithms behavior can be described by the following steps \citep{saxena:17}:

\begin{enumerate}
  \item Start with a singular cluster containing $N$ documents.
  
  \item Split the cluster that yields the two components with the largest inter-cluster distance.

  \item Repeat step 2 until the clustering is satisfactory.
    
\end{enumerate}

Divisive algorithms behavior can be described by the following steps \citep{saxena:17}:

\begin{enumerate}
  \item Start with a singular cluster containing $N$ documents.
  
  \item Split the cluster that yields the two components with the largest inter-cluster distance.

  \item Repeat step 2 until the clustering is satisfactory.
    
\end{enumerate}

For better understanding, according to \citet{Hansen:1997}, the distance matrix is defined as ${D = (d_{kl})}$, an ${N \times N}$ matrix, where ${N}$ is the number of clusters and ${d_{k \times l}}$ is the distance between the clusters ${k}$ and ${l}$. The distances within the matrix generally satisfy the following conditions: ${d_{kl} \geq 0}$, ${d_{kk} = 0}$ and ${d_{kl} = d_{lk}}$, consider for ${k,l = 1, 2,...,N}$. Instead of using a distance matrix, a similarity matrix can be used, so storing the similarity between the clusters. In this case, in the step 2 the algorithm should search for the maximum similarity.

\subsection{Incremental Clustering}

Most of the clustering algorithms in the literature do not deal with dynamic data, i.e., they require the entire database to be available in order to perform the clustering. With the advent of the Web, where new information emerges dynamically, such algorithms may not be effective \citep{Sahoo:2006}. In order to address that challenges, incremental clustering algorithms, capable of working with data streams, have been proposed. 

In this context, data streams are described as ordered sequences of data that are continuously generated \citep{Guha:2003, Silva:2013}. As formally defined by \citet{Silva:2013}, a data stream $S$ is a sequence of data objects $d_1, d_2, \dots, d_N$, where $S = \{d_i\}_{i=1}^N$, and this sequence can be infinite ($N \rightarrow \infty$).

In this scenario, it can be observed that the algorithm does not have a complete dataset available at the beginning of the clustering process, thus a linear reading of the data is necessary \citep{Guha:2003}; the data are read in sequence as they become available in order to avoid rereading data that has already been processed. Such restrictions make working with dynamic data more complex than work with static data \citep{PhridviRaj:2014}. Thus, the main characteristic of incremental clustering is that the algorithm is able to adapt its clustering structures by adding or removing a data object, and thus, regrouping of all the data is not required \citep{Garcia:2010a}. Depending on the algorithm used, after many increments, this total regrouping of data would be valid to improve the quality of the groups. Further, these algorithms are used in contexts with a large volume of data and it is not possible to keep them in the memory, and also in situations in which the entire data are not stored, but only their summaries are stored \citep{Guha:2003}.

According to \citet{Silva:2013}, in order to develop algorithms that mine data streams, the following conditions should be considered:

\begin{enumerate}
  \item New data will be continuously obtained;

  \item There is no defined order in which the data are processed;

  \item The data stream can be unlimited;

  \item The data are discarded after being processed, thus avoiding being reread;

  \item The process for data generation is unknown and non-stationary.
\end{enumerate}


\section{Systematic Review Process}
\label{sec:pesquisa_sistematizada}

This section describes the systematic review process used to collect recently published works related to incremental hierarchical text clustering techniques. This study aimed to obtain recent studies on clustering techniques that have some specific characteristics, which are listed with their corresponding key words below:

\begin{itemize}
  \item \textbf{Clustering techniques}: cluster, clustering;
  
  \item \textbf{Hierarchical clustering}: hierarchical, hierarchy;
  
  \item \textbf{Textual document clustering}: document, text, textual, news, topic;
  
  \item \textbf{Incremental clustering}: data stream, datastream, dataflow, data flow, incremental, dynamic.
\end{itemize}

Based on these terms, a search was created, as shown in Figure \ref{fig_pesquisa_sistematizada}, and applied to the ACM, Science Direct, Springer Link and IEEE Xplore digital libraries. The attributes ``Title", ``Abstract", and ``Keywords" were used in the search fields. Papers from January 2010 to April 2018 were considered.

\begin{figure}[ht!]
  \setlength{\abovecaptionskip}{0pt}
  \begin{framed}
  \centering

    ( ``cluster" OR ``clustering" ) \\
    AND \\
    ( ``hierarchical" OR ``hierarchy" ) \\ 
    AND \\ 
    ( ``document" OR ``text" OR ``textual" OR ``news" OR ``topic") \\
    AND \\ 
    ( ``incremental" OR ``dynamic" OR ``datastream" OR ``data stream" OR ``dataflow" OR ``data flow" )

  \end{framed}
  \caption{Query used in the systematic review process.}
  \label{fig_pesquisa_sistematizada}
\end{figure}

The objective of this review was to search for studies that developed new techniques or approaches for hierarchical clustering of texts in dynamic environments. However, during the search, some manuscripts related to the applications of these techniques were also found and are described here.

When the search was complete, a total of 169 published papers were obtained: 17 from the Science Direct library, 50 from ACM, 64 from Springer Link and 38 from IEEE Xplore. Among these papers, 10 were duplicate studies, found in more than one library. After reading the papers, a subset of 30 papers related to the topic was selected; of these, 17 papers were related to the development of a new clustering technique, and the other 14 paper were concerned with the application of the techniques in a certain context. The other works were not selected because, despite meeting the search made, they did not meet one of the topics: hierarchical, incremental and text clustering. Table \ref{tab_resultado_pesquisa_sistematizada_1} shows the number of papers obtained per digital library, and Table \ref{tab_resultado_pesquisa_sistematizada_2} shows the number of selected published papers per purpose.

{
\renewcommand{\arraystretch}{2.0}
\begin{table}[!ht!tp]
  \centering
  \caption{Published papers obtained through the systematic review process.}
  \setlength{\tabcolsep}{8pt}
  \scriptsize

  \begin{tabular}{lcc}
  
        \textbf{Digital Library}          & \textbf{Number of Papers}   \\ \hline 
                                    
        Science Direct              & 17           \\ \hline

Springer Link              & 64           \\ \hline

        ACM                         & 50         \\ \hline

        IEEE Xplore                 & 38          \\ \hline
    
    \textbf{Total} & \textbf{169} \\ \hline
  
    \end{tabular}

    \label{tab_resultado_pesquisa_sistematizada_1}
\end{table}
}%

{
\renewcommand{\arraystretch}{2.0}
\begin{table}[!ht!tp]
  \centering
  \caption{Published papers per purpose.}
  \setlength{\tabcolsep}{8pt}
  \scriptsize

  \begin{tabular}{lcc}
  
        \textbf{Paper purpose}            & \textbf{Number of Papers}   \\ \hline 

        New clustering techniques        & 17          \\ \hline

        Applications                     & 14           \\ \hline
        
        \textbf{Total}                   & \textbf{31} \\ \hline
  
    \end{tabular}

    \label{tab_resultado_pesquisa_sistematizada_2}
\end{table}
}%

\begin{figure}[ht!]
  \centering
  \includegraphics[width=9cm]{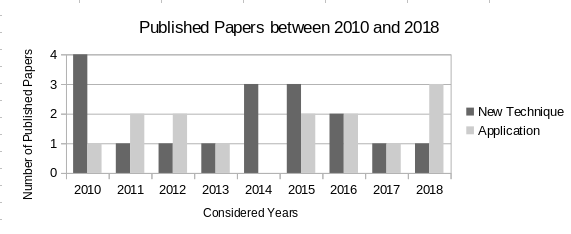}
  \caption{Chart showing the obtained published papers per year.}
  \label{fig_grafico_trab_relacionados_anos}
\end{figure}

Figure \ref{fig_grafico_trab_relacionados_anos} shows a chart with the total number of papers selected for each year in the period considered. It is important to highlight that this search was conducted in the digital libraries in April 2018, and hence, if the search is repeated in the future, other papers may be found.

After obtaining and selecting the published papers, they were studied and compared. These comparisons were performed only for those studies whose objective was the development of new clustering techniques; from these studies, the following information was collected:

\begin{enumerate}

  \item Similarity and distance measures used;

  \item Databases used in the experiments;

  \item Metrics used to evaluate the algorithms;

  \item Baseline algorithms used in the comparisons.
\end{enumerate}

The papers obtained by this systematic review are briefly described in Section \ref{sec:trabalhos_encontrados}, and, then, the comparisons of these aspects are presented.


\section{Literature Survey}
\label{sec:trabalhos_encontrados}

This section presents a brief description of the related works found using the systematic review process. As described in Section \ref{sec:pesquisa_sistematizada}, these papers are categorized by their purpose: development of new techniques (develops new clustering techniques or approaches) and application of techniques (applies clustering techniques in a specific domain). It is important to note that the ``application" studies did not necessarily use the techniques proposed in the ``development of new techniques" studies.

Table \ref{sumario_trabalhos_encontrados} summarizes the articles that applied techniques and Table \ref{sumario_trabalhos_encontrados_tenicas} summarizes the articles that created new techniques. Sections \ref{sec:medidas_similaridade}, \ref{sec:bases_de_dados}, \ref{sec:metricas} and \ref{sec:algoritmos_comparados} present additional details about some characteristics of these published studies. These sections apply only to the ``development of new techniques" works.

\subsection{Clustering Technique Applications}
\label{sec:trabalhos_encontrados_aplicacoes}

In this section, some applications of hierarchical clustering of documents in dynamic environments are presented. The main focus of these works was not the development of a new clustering techniques but its application in a specific domain.

{
\renewcommand{\arraystretch}{2.0}
\linespread{1.00}
\begin{table*}[!ht] 
  \caption{Summarization of the Application studied papers.}
  \centering
  \begin{small}
  \scriptsize
  \begin{tabular}{p{6cm}p{2cm}p{2cm}p{1.9cm}}

\textbf{Reference}  
    & \textbf{Year} 
    & \textbf{Libraries}   \\ \hline

\citet{Wang:2010}  
    & 2010  
    & ACM, Springer \\ \hline

  \citet{Gao:2011} 
    & 2011  
    & ACM, IEEE \\ \hline

 \citet{Luo:2011} 
    & 2011  
    & SD  \\ \hline

  \citet{Lin:2012} 
    & 2012     
    & ACM, IEEE \\ \hline

  \citet{Lu:2012}  
    & 2012  
    & IEEE  \\ \hline

  \citet{gunaratna:15}  
    & 2015  
    & ACM  \\ \hline

  \citet{Mythily:2015}  
    & 2015  
    & SD  \\ \hline

  \citet{zhiyi:2015}  
    & 2015  
    & IEEE  \\ \hline

  \citet{hoxha:2016}  
    & 2016  
    & SD  \\ \hline
    
\citet{protasiewicz:2016}  
    & 2016  
    & IEEE  \\ \hline

  \citet{popat:17}  
    & 2017
    & IEEE  \\ \hline

   \citet{amelin:18}  
    & 2018
    & SD  \\ \hline
    
    \citet{huang:18}  
    & 2018
    & Springer  \\ \hline
    
    \citet{sangaiah:18}  
    & 2018
    & Springer  \\ \hline
    
  \end{tabular}
  \end{small}

  \label{sumario_trabalhos_encontrados}
\end{table*}
}

\citet{Wang:2010} used a hierarchical clustering method that focuses on the summary of documents and updates the summaries as soon as a new document is added. The COBWEB \citep{Fisher:1987} algorithm was used, and was adapted for a sentence hierarchy. They concluded that the proposed algorithm was able to produce higher quality summaries than the other techniques used in the experiments.

The study of \citet{Gao:2011} focused on a different aspect: the relationship between the topics. They used the hierarchical Dirichlet process (HDP) model to group documents into topics; then, events in these topics in a dynamic environment are monitored. Examples of such events include a junction between two topics to form a single topic or the division of one topic resulting in two topics. A method to view the evolution of the topics over time was also developed.

In the study of \citet{Luo:2011} a method to categorize the eligibility criteria for patients in clinical trials was proposed. The hierarchical agglomerative clustering algorithm (HAC) was applied on the semantic feature matrix to obtain the semantic categories. The categories obtained represent characteristics, such as age groups. Supervised machine learning classifiers and hierarchical clustering were combined to yield a semi-supervised learning approach; the final objective was to associate the eligibility criteria with the actual patient data.

\citet{Lin:2012} proposed the application of incremental hierarchical clustering for documents from the Legislative Yuan Library website. In the presented method, only the named entities (like people, locations and organizations) \citep{Mikheev:1999} were extracted for usage in the clustering. The clustering method consists of two steps: (1) a hierarchical clustering algorithm is used to identify the ideal number of clusters, and then, (2) a k-means algorithm \citep{Macqueen:1967} is used to cluster the documents. The method was compared with categorizations performed manually by specialists, and the proposed method demonstrated a satisfactory success rate.

\citet{Lu:2012} developed a framework for detecting hot topics in the news on the Internet. In this framework, an incremental clustering algorithm is used to identify the candidate topics from the news in a time window. Then, an algorithm based on K-Nearest Neighbor and Single Pass \citep{Rijsbergen:1979} algorithms is used to select the topics from the candidates. In the final step, an algorithm to generate the descriptors for the identified topics was also presented. The authors reported that the algorithm achieved good performance in comparison with other algorithms.

In the work of \citet{Mythily:2015} was proposed a system whose objective is the grouping of news from data streams. In this system news are read from an RSS stream and, after a pre-processing of the data, the news is grouped according to the event they describe. The clustering performed in the developed system uses algorithms known as "segment-wise distributional clustering" and the k-means algorithm.

\citet{gunaratna:15} adapted a clustering algorithm called Cobweb \citep{Fisher:1987} that aimed to identify conceptually similar groups of facts of an RDF entity description identity. They proposed a new approach called FACeted Entity Summarization (FACES) to add orthogonal semantic groups of facts to diversify the summaries and verify the effectiveness of this approach according to traditional techniques. The technique developed was compared with other techniques of the state of art (RELIN, RELINM and SUMMARUM). The authors created a gold standard for the evaluation due to unavailability of the evaluation data or RELIN. The evaluation was tested using the gold standard and user preference. Their approach showed superior results, improving the summary quality.

 \citet{zhiyi:2015} used hierarchical clustering to identify bursty events by users reports on Twitter. They purposed a BBW (Basic -Burst Weight) based on the Time Window to extract bursty words and combined it with hierarchical clustering that aimed to discover events. 

\citet{hoxha:2016} presented an application for learning taxonomies relations from text using background knowledge. They applied an hierarchical agglomerative clustering technique to group relative concepts under the same cluster. Tests were made with real world text collections, using data from clinical trial eligibility criteria descriptions and documents describing drug-drug interactions. On their result, they concluded the application presents higher concept coverage and higher accuracy of learned taxonomic relations than existing dynamic pruning and the state-of-art taxonomy learning methods.

In the \citet{protasiewicz:2016} study, it was proposed a hybrid knowledge-based framework for author name disambiguation that aims to identify the author of some document or even create a new profile if it still doesn't exist. Their experiments occurred in three steps: (1) tests of the rule-based algorithm to verify its accuracy in a collection with around 19000 references disambiguated; (2) tests with hierarchical agglomerative clustering algorithm with the similarity calculated by the heuristics rules, witch tests resulted in high precisions but low recall; (3) classifiers were used with the hierarchical agglomerative clustering algorithm for estimating a similarity of clusters. It was tested with four linkage functions, four classifiers and five stop criteria.

\citet{popat:17} applied the HSC algorithm for measuring the similarity between data text documents. The tests were performed with datasets with thousands of documents. They were pre-processed and then, clustered. The similarity was measured by a new cosine method, because of the need of calculating frequencies in an efficient way and it fits well with comparison vector representation of two documents. The results were compared with other techniques and the HSC performed well compared to other clustering techniques.

In the work of \citet{amelin:18}, they aimed study the patterning of the evolution of writing style. It was studied the evolution of some book collections to view the capacity of identifying the style development over time. They proposed a new clustering procedure to manage very large data sets. The proposed method works in two steps: In the first step, a partition algorithm is applied to form the small "pre-clusters". An algorithm like k-means can be used, so the number of clusters can be previously determined by a cluster validation technique. The formed clusters are applied into a hierarchical agglomerative clustering combined with the clusters previously formed. The method was able to identify the author of a set of books automatically by the writing pattern.

\citet{huang:18} proposed a framework that aimed to capture important moments through analyzing Twitter streams. The event summarization consists of three key components: participant detection, sub-event detection and sumary tweet extraction. In the participant detection phase, the authors compare four different approaches: Exact Match, TweetUpdate, IncNameHAC and NameHAC. The best performace was reached by NameHAC, a hierarchical clustering method. Its incremental version, the  IncNameHAC does not perform so well. The participant detection key aims to identify the important entities in the stream that play a significant role in shaping the event progress; The sub-event detection detects the important moments and the summary generation module takes the imput of sets of tweets aiming to generate a high-quality textual summary.

\citet{sangaiah:18} used incremental k-means clustering to provide conjectural navigation and browsing techniques to arabic text documents. The authors consider arabic language challenging for its orthographic variations, different ways to wrie certain combinations of characters, a complex morphology and the words are often ambiguous. Their approach consists in two stages, document preprocessing and clustering. The first one includes prepare the text documents and apply a term weighting method. The clustering stage was executed with partitional and hierarchiqual algorithms and showed the effectivenes of unsupervised leaning, semi-supervised leaning, and semi-supervised learning with dimensionality reduction algorithms by using k-means, incremental k-means, Threshold + k-means and k-means with dimensionality reduction.
\subsection{New incremental hierarchical clustering techniques}
\label{sec:trabalhos_encontrados_tecnicas}

Here are described the works that developed a new technique to classify
text data.

{
\renewcommand{\arraystretch}{2.0}
\linespread{1.00}
\begin{table*}[!ht] 
  \caption{Summarization of the New Technique studied papers.}
  \centering
  \begin{small}
  \scriptsize
  \begin{tabular}{p{8cm}p{2cm}p{1.9cm}}

\textbf{Reference}  
    & \textbf{Year} 
    & \textbf{Libraries}   \\ \hline

\citet{CorreaMorris:2010}
    & 2010  
    & SD, ACM \\ \hline

  \citet{Garcia:2010a} 
    & 2010  
    & SD, ACM \\ \hline

  \citet{Garcia:2010b} 
    & 2010  
    & ACM, Springer \\ \hline

  \citet{LWang:2010} 
    & 2010  
    & IEEE  \\ \hline

\citet{Huang:2011} 
    & 2011  
    & IEEE  \\ \hline

 \citet{Marcacini:2012} 
    & 2012  
    & ACM, Springer \\ \hline

 \citet{Marcacini:2013} 
    & 2013  
    & ACM \\ \hline
    
  \citet{wang:13} 
    & 2013  
    & ACM \\ \hline

\citet{Cai:2014} 
    & 2014
    & SD  \\ \hline

 \citet{cui:14} 
    & 2014
    & IEEE  \\ \hline

  \citet{Sinoara:2014} 
    & 2014  
    & ACM \\ \hline

  \citet{Peng:2015}  
    & 2015  
    & SD  \\ \hline

 \citet{Wang:2015}  
    & 2015  
    & SD  \\ \hline

\citet{irfan:16}  
    & 2016
    & IEEE, Springer  \\ \hline

\citet{khalilian:16}   
    & 2016
    & Springer  \\ \hline

  \citet{kobren:17}  
    & 2017  
    & ACM \\ \hline
    
     \citet{sutanto:18}
    & 2018  
    & Springer \\ \hline

  \end{tabular}
  \end{small}

  \label{sumario_trabalhos_encontrados_tenicas}
\end{table*}
}

\citet{Garcia:2010a} presented a framework for incremental hierarchical document clustering in which a graph is used to store each level of the hierarchy. In this framework two graphs are created for each level of the hierarchy, where the vertices are the clusters and edges the similarities between them. The first one is the $\beta$-similarity graph, where the connected vertices are those whose similarity and greater than a parameter $\beta$. From this graph is created max-S graph, where each vertex is connected by his edge of greatest weight (similarity). Thus, the clusters for the level are created from a vertex cover in max-S graph and these will be the vertices of the next level of the hierarchy.
Based on this framework, two algorithms were presented: Dynamic Hierarchical Compact (DHC) and Dynamic Hierarchical Star (DHS). The DHC algorithm constructs disjoint clusters, while the DHS algorithm allows for overlap among clusters. The results demonstrated that the proposed algorithms have superior performance than the Unweighted Pair Group Method with Arithmetic Mean (UPGMA) \citep{Jain:1988} and the Bisecting K-Means (BKM) \citep{Karypis:2000} algorithms. According to \citet{Garcia:2010a}, the new techniques can create easily navigable hierarchies because they aim to balance their width with respect to their depth.

\citet{Garcia:2010b} proposed an improvement in the DHC algorithm \citep{Garcia:2010a}. Based on the observation that the updates in the structures used in the algorithm require various similarity calculations between the clusters, a new step was added to a local selection of the most relevant terms for the clusters, thus resulting in the reduced dimensionality of these data. Hence, a considerable computational cost reduction in the DHC algorithm was obtained with minimal loss of quality in the built hierarchies. Further, a new metric was presented to evaluate the quality of the created hierarchies. As explained by \citet{Garcia:2010b}, this metric aims to measure the navigation cost of finding the desired topic, and is useful in experiments to generate hierarchies with good navigability.

In the study of \citet{CorreaMorris:2010}, an algorithm that calculates nested partitions was proposed; this method was called Incremental Nested Partition Method (INPM). It uses different clustering criteria at each level of the hierarchy and structures the documents into graphs such that the mathematical properties of this structure can be used. Given the documents to be grouped, each document is represented as a vertex in the graph and two vertices are connected if the similarity between them is greater than a threshold. This graph is constructed according to the similarity measure and clustering criterion used and the groups are obtained from the connected components of the graph.
The main strengths of this algorithm is that it is not sensitive to the order in which the documents are added, and the levels of the hierarchy can be obtained independently simply building the similarity graph of the level, using the level clustering criteria and finding the connected components. Another interesting property presented by the authors is the ability to calculate the number of partitions at a level without executing the entire process.

\citet{LWang:2010} proposed an algorithm based on dynamic Indexing Trees, called Multi-Representation Indexing Trees (MRIT). This tree is used to represent the document hierarchy and to insert a new document in this structure, the algorithm runs a search into the tree from the root, exploring only the nodes with similarity greater than a threshold, until reaching terminal nodes, when the most similar leaf is selected. Thus, the path to this leaf is scanned again from the root to find the correct node into which the document will be inserted. The method proposed by \citet{LWang:2010} differs from the traditional Indexing Tree in the cluster representation. In the original tree, each cluster is represented by its centroid, and this is used in calculations of similarity; in the MRIT each cluster has several representative points, and these points are considered in the calculation of similarity and the cluster threshold.
Regarding the quality, better results were obtained using the MRIT, in comparison to original Indexing Tree. With regard to efficiency, the use of traditional Indexing Tree has cost of O(n) cost while MRIT has O (n * k), where k is the maximum number of representative points in a cluster. In this case, as k << n, the efficiency of both approaches is fairly close. The advantage of the presented technique is the insensibility to the order of insertion of the documents and the avoiding the decentralization of the clustering center. The weak point of this work is the lack of comparisons with other algorithms, and also the tests with other datasets.

The paper of \citet{Huang:2011} presented a strategy for extracting news topics that consists of two steps: retrospective clustering, which uses the agglomerative hierarchical clustering algorithm to group the news already found in the database (past news), and online clustering, which uses the Single Pass algorithm \citep{Rijsbergen:1979} to add the latest news to the hierarchy of topics. In the second step, the news is organized into 24-hour windows, and clustered into micro-clusters that are later inserted into the hierarchy in existing clusters or in new ones. The study used the extraction of named entities to help in the clustering tasks, and assigned a greater weight to these terms. In the presented results, the strategy of using named entities proved to be more effective than the strategy of considering the words from the title to have greater weight.
 
\citet{Marcacini:2012} presented an approach of using consensus clustering for the Buckshot algorithm \citep{Cutting:1992}. The generation of the initial model is considered as a critical phase for incremental clustering algorithms because the errors that occur in this phase can be propagated to the other documents throughout the clustering process. In an attempt to solve this problem, the Buckshot Consensus Clustering algorithm was presented in this study. In this approach, a first set of data is obtained and is performed offline clustering of these data using different algorithms. The obtained results are integrated into a single solution, thus creating a more robust initial model that is less sensitive to degradation. After that, the algorithm begin receiving new data and run the incremental clustering. According to the authors, this method demonstrated good results when compared with non-incremental algorithms and algorithms that do not use consensus clustering.

\citet{Marcacini:2013} proposed a method that uses privileged information for document clustering. Privileged information is additional information that is specific to a domain; sometimes it's difficult or expensive to obtain such information and is available for only a subset of data. Generally, the domain specialists provide such information. The technique presented, referred to as LUPI-Based Incremental Hierarchical Clustering (LIHC), uses the subset containing this privileged information and applies various clustering algorithms to it; consensus clustering is used to generate the initial partitioning data model. Based on this model, the other documents are incrementally inserted into the clusters, and the structure is updated. The proposed method achieves better quality of the hierarchies created than the Buckshot algorithm.

In the work of\citet{wang:13}, it was proposed an evolutionary multi-branch tree clustering method for understanding topics in a text stream. They aimed to provide a coherent view of content transitions, because existing works didn't provide interpretable topics results once the most of the topic trees are not binary. The algorithm developed was compared with evolutionary hierarchical clustering \citep{chakrabarti:06}, that according the authors, the tests showed that the developed algorithm outperforms the baseline algorithm in all aspects; were evaluated the efficiency of the algorithms, the effectiveness of constraint model, tree likelihood and tree smoothness.

In the work of \citet{Cai:2014}, a hierarchical clustering framework that mainly focuses on obtaining meaningful hierarchical levels was proposed. This framework can be used with any similarity measure that complies with certain restrictions as detailed in the study, such as triangle inequality with a relaxation factor. The framework was also adapted to deal with dynamic contexts. In the conducted experiments, the framework was combined with the k-means \citep{Macqueen:1967} and k-medians \citep{Jain:1988} algorithms, and demonstrated good results when compared with other algorithms.

\citet{cui:14} created a new method about organization of text corpora by using hierarchical approach. Their tests focused in the quality of the clusters and the ability of users to explore and identify the evolving topics. To provide the quality and a better understand  for the users, they used evolving topic trees to organize documents at different times. The evolutionary topic tree was built by using the multi-branch tree clustering algorithm, that allowed to use this approach hierarchically and incrementally.

Using the LIHC algorithm \citep{Marcacini:2013}, the study of \citet{Sinoara:2014} proposed the use of named entities contained in the text as privileged information. For the considered datasets, tools were utilized for the extraction of the entities from one part of the data, and these entities were used to generate the initial clustering model. In the experiments conducted, the results did not show a significant improvement in the hierarchies quality with the use of named entities. However, the authors concluded that the use of named entities contributed to the generation of more intelligible cluster descriptors. They also observed that the quality of the extracted entities can directly affect the quality of the obtained clusters.

The study of \citet{Peng:2015} presented an agglomerative method for hierarchical clustering of documents, referred to as ICHTC-CF (Incremental Conceptual Hierarchical Text Clustering Using CFu-tree). In this method, the CFu-Tree is used to represent the clustering hierarchy and is shown to be effective in the context of incremental clustering. An important issue of this method is the absence of a parameter that establishes the maximum number of clusters (such as the K parameter in the K-means algorithm), because choosing a value for this parameter can be a challenge. Thus, the Comparison Variation (CV) measure was proposed to determine whether the two clusters should be united during the process, which guarantees the effectiveness of the method. According to the authors, this technique showed better results than the algorithms K-means \citep{Macqueen:1967}, CLIQUE \citep{Agrawal:1998}, Single Linkage \citep{Jain:1988}, and Complete Linkage \citep{Jain:1988}, and an increase in the effectiveness with an increase in the number of terms (features).

The study of \citet{Wang:2015} presents a Hierarchical Dirichlet process (HDP) to cluster documents. A different feature of the proposed method is the ability to work in a partial supervised way, working with a mixture of documents with and without labels. If there is an initial set of classified documents, the algorithm is able to use this information in order to generate a more robust model, and to adapt it incrementally with new documents. If there is no labeled document, it works as a traditional unsupervised HDP. Another feature of this method is a parameter that represents the depreciation rate of the model, which indicates how fast the model will forget old information to better fit the new ones. This feature avoids the unlimited growth of the model, avoiding the loss of efficiency of the method. The algorithm also has another parameter that represents the probability of a new group being created when inserting a new document. In the experiments, the proposed method was compared with traditional supervised algorithms, like Support Vector Machine (SVM) \citep{Cortes:1995} and Naive Bayes (NB) \citep{Hassan:2011}, presenting satisfactory results.

In the study of \citet{irfan:16}, they aimed to find an incremental solution for evolving an existent taxonomy by a novel algorithm. The algorithm was called Taxonomy Incremental Evolution (TIE). When a new document is detected, the closest cluster is identified by calculating the similarity between the document and the current clusters in a multidimensional space. 
Therefore, to reorganize the existing taxonomy, it's necessary to check the quality of each cluster and then, decide if it will be necessary to relabel or merge clusters or restructure the data. The goal is to evolve the version of the existing taxonomy. The method was compared with the technique of regeneration and it showed to be more efficient in time and quality.

\citet{khalilian:16} proposed a new technique based on STREAM and ConStream methods. The new framework Divide and Conques Data Stream (DCSTREAM) technique was proposed to cluster huge datasets using data stream clustering and is partitioned into online and offline components. The online module consists on three components: subsets generator, that creates one dimention vector for clustering by the k-means algorithm; micro-clusters generator; and the split and merge component and the clusters formed are stored on statistical databse. The offline component provides a hierarchical structure for clusters by statistical databse.

In the study of \citet{kobren:17}, they proposed a novel clustering algorithm that aims to be well scalable with a large number of clusters, not only with a large number of data points. The hierarchical characteristic of the technique was essential to this objective. The algorithm is called Purity Enhancing Rotations for Cluster Hierarchies (PERCH), the method consists in a greedy incremental tree construction. After the greedy algorithm, if a masking is detected, they execute a rotate procedure in four steps to solve it. After being compared with other algorithms in several datasets (some textual), the results showed that Perch is competitive with other state-of-art algorithms for small benchmarks, but for large datasets, it produces trees with highest dendogram purity and best F\textsubscript{1}-measure results.

\citet{sutanto:18} focused on extraction of insights from a large volume of text documents. They developed the an incremental clustering algorithm named Fine-Gfranted Document Clustering via Ranking (FGCR), focused on overcoming the computational complexity and the difficult of the high dimensionality of data by using loci and relevante clusters, avoiding the need of scanning all of the data to make clustering decisions. The asymptotic complexity of FGCR is O(tmN), where t be the querying time to form S (the relevant documents set), m be a constant defined by the size of S and N the number of documents. The new technique shown suitable to partition a large text collection with numerous topics such as social media data.


\subsection{Similarity and Distance Measures}
\label{sec:medidas_similaridade}

Clustering techniques use similarity or dissimilarity measures to determine whether elements must be inserted into the same cluster \citep{Cherkassky:2007}, and these measures impact directly in the clustering process quality. The choice of the measure to be used requires careful consideration because various types and ranges of data exist \citep{Jain:1999}, and the context of the problem is also a factor that should be considered \citep{Huang:2008}. In order to measure the similarity between two elements, similarity measures and dissimilarity measures can be used because these two measures are often interchangeable \citep{Xu:2009}. In the case of similarity measures, as their value increases, the similarity between the two documents increases and the probability of them being inserted in the same cluster also increases. On the other hand, in the case of dissimilarity measures, as their value increases, the similarity between the documents decreases and the probability of them being inserted in the same cluster also decreases.

Table \ref{sumario_medidas_similaridades} presents the most commonly used measures in the studied works. Other measures are summarized by \citet{Xu:2005} and \citet{Huang:2008}.

{
\renewcommand{\arraystretch}{2.0}
\linespread{1.0}
\begin{table*}[!ht]
  \centering
  \caption{Similarity measures used in the studied works.}
  \begin{small}
  \setlength{\tabcolsep}{3pt}
  \scriptsize
  \begin{tabular}{p{5.4cm}p{1.5cm}p{4.6cm}}
  
  \textbf{Measure}  & \textbf{Type} & \textbf{Studies where applied}  \\ \hline

  \textbf{Cosine Similarity} \newline \citep{Xu:2005, Huang:2008}
    & Similarity 
    & \citet{Garcia:2010a} \newline
      \citet{Garcia:2010b} \newline
      \citet{Wang:2010} \newline
      \citet{Huang:2011} \newline
       \citet{Marcacini:2012} \newline
      \citet{wang:13} \newline
      \citet{cui:14} \newline
      \citet{Sinoara:2014} \newline
       \citet{protasiewicz:2016} \newline
      \citet{irfan:16} \newline
      \citet{popat:17}
      \newline
      \citet{huang:18}
      \newline
      \citet{sangaiah:18}
    \\ \hline
  
  \textbf{Euclidian Distance} \newline \citep{Xu:2005, Huang:2008}
    & Distance 
    & \citet{Cai:2014} \newline
      \citet{Peng:2015} \newline
      \citet{popat:17}\\ \hline

  \textbf{Weighted sum of $\delta$-kernels} \newline \citep{Scholkopf:2001, Taylor:2004}  
    & Similarity 
    & \citet{CorreaMorris:2010}      \\ \hline
  
  \textbf{Polynomial kernel of degree 1} \newline \citep{Scholkopf:2001, Taylor:2004} 
    & Similarity 
    & \citet{CorreaMorris:2010}      \\ \hline

  \textbf{Prediction Measure} \newline \citep{wang:13} 
    & Similarity 
    &  \citet{wang:13}      \\ \hline

  \textbf{Distributional Similarity} \newline \citep{fan:07} 
    & Similarity 
    &  \citet{hoxha:2016}      \\ \hline

  \textbf{Spearman's Correlation Distance} \newline 
  \citep{spearman:04} 
    & Distance 
    &  \citet{amelin:18}      \\ \hline

 \textbf{Canberra Type Distance} \newline 
  \citep{lance:66} 
    & Distance 
    &  \citet{amelin:18}      \\ \hline

  \end{tabular}
  \end{small}

  \label{sumario_medidas_similaridades}
\end{table*}
}

The sections \ref{sec:similaridade_cosseno} and \ref{sec:distancia_euclidiana} describe in details two of the most commonly employed measures and then, the section \ref{sec:other_similarities} will describe the other less used measures. For this purpose, $d_i$ and $d_j$ are used to represent two different documents, $x_i$ and $x_j$ are their respective term vectors, $p$ denotes the number of dimensions of these vectors, and $\alpha$ is the angle formed by these two vectors.

\subsubsection{Cosine Similarity}
\label{sec:similaridade_cosseno}

Cosine similarity is a commonly used measure for clustering textual documents. In this measure, the size of the vectors $x_i$ and $x_j$ is not considered; instead, the directions of the vectors and the angle formed by them are considered. Thus, in the context of text clustering, this means that documents composed of the same terms (and the same proportion among them) but having different total number of terms would be considered similar \citep{Huang:2008}. Equation \ref{eq_distancia_cosseno} shows the calculation of this similarity measure, which corresponds to the cosine of the angle formed by the two vectors ($x_i$ and $x_j$) \citep{Xu:2005}.

  \begin{equation}
    D_{cos}(d_{i},d_{j}) = cos\:(\theta) = \frac{ x_{i} \cdot x_{j} }{ \| x_{i} \| \cdot \| x_{j} \| }
    \label{eq_distancia_cosseno}
  \end{equation}

\subsubsection{Euclidean Distance}
\label{sec:distancia_euclidiana}

The most popular distance measure, used mainly for continuous values, is the Euclidean distance, shown in Equation \ref{eq_distancia_euclidiana}. This measure is intuitive because it is used to calculate the distance between objects in space (geometric distance) \citep{Jain:1999}. 

  \begin{equation}
    D_{euc}(d_{i},d_{j}) = \sqrt{ \sum_{k = 1}^{d} (x_{i,k} - x_{j,k})^2 }
    \label{eq_distancia_euclidiana}
  \end{equation}

\subsubsection{Other similarities}
\label{sec:other_similarities}

On the studies of \citet{CorreaMorris:2010} were used the weighted sum of $\delta$-kernels and polynomial kernel of degree 1. Given the complexity of the existence of polynomial classifiers, a weighted sum of a kernel function reduces the number of computations by the dual space representation to reduce the number of computations required \citep{Scholkopf:2001, Taylor:2004}. The degree 1 of a kernel leads to a linear separation of the data. 
  
The Prediction Measure is a probability-based measure used by \citet{wang:13}, this measure outperformed the cosine measure on the texts. 
\citet{hoxha:2016} used the Distributional Similarity, that is used in conjunction with a knowledge source to compute similarity. Distributional based methods are a Semantic Similarity measure that help to better understand a textual resource. Distributional methods generally apply syntactic approaches to find out the similarity regarding predicate-argument relations \citep{hoxha:2016}. 

\citet{amelin:18} used and compared the Spearman's Correlation Distance and Canberra Type Distance to measure the similarity of two distributions obtained by Vector Space Model. The Spearman Correlation is a rank-based, nonparametric and size-independent technique for evaluating the degree of linear association or correlation between two independent variables.


\subsection{Datasets}
\label{sec:bases_de_dados}

A distinguishing factor among the studied papers is the datasets used. Table \ref{tab_datasets_trabalhos_relacionados} summarizes the datasets that were used in at least two studies from the collected ones and the sources that they were collected. The column dataset describes the name of the collection used and the column source describes where the the collection was obtained. It is important to highlight that some studies used only a subset of the dataset.

{
\renewcommand{\arraystretch}{2.0}
\linespread{1.0}
\begin{table*}[!htp]
\centering
\caption{Datasets used in at least two studies.}
\setlength{\tabcolsep}{3pt}
\scriptsize
\begin{tabularx}{\textwidth}{XXp{1.2cm}p{1.4cm}p{1cm}p{4.6cm}}

\textbf{\newline Dataset} 
  & \textbf{\newline Source}  
  & \textbf{Num. \newline Doc.} 
  & \textbf{Num. \newline Terms} 
  & \textbf{Num. \newline Classes} 
  & \textbf{\newline Studies where applied}  \\ \hline 
  
\textbf{Afp} 
  & TREC-5      & 695   & 12575   & 25
  &  \citet{Garcia:2010a} \newline
    \citet{CorreaMorris:2010}   \\ \hline

\textbf{Eln} 
  & TREC-4      & 5829    & 84344   & 50 
  &  \citet{Garcia:2010a} \newline
     \citet{Garcia:2010b}  \\ \hline

\textbf{Tdt}          
  & TDT2        & 9824    & 55112   & 193
  & \citet{Garcia:2010a} \newline
   \citet{Garcia:2010b} \newline
    \citet{CorreaMorris:2010}  \\ \hline

\textbf{Reu}          
  & Reuters-21578 & 10369 & 35297   & 120
  & \citet{Garcia:2010a} \newline
    \citet{Garcia:2010b}  \\ \hline

\textbf{Ohscal} 
  & Ohsumed     & 9200    & 13512   & 12 
  &  \citet{Garcia:2010a} \newline
     \citet{Garcia:2010b}  \\ \hline

\textbf{Reviews}      
  & San Jose \newline Mercury & 4069    & 22927   & 5 
  & \citet{Garcia:2010a} \newline
     \citet{Marcacini:2012}  \\ \hline

\textbf{Hitech}       
  & San Jose \newline Mercury & 2301    & 12942   & 6
  & \citet{Garcia:2010a} \newline 
     \citet{Garcia:2010b} \newline 
     \citet{Marcacini:2012}  \\ \hline

\textbf{20ng}       
  & 20ng        & 18808 & 45434   & 20 
  &   \citet{Marcacini:2012} \newline
      \citet{Marcacini:2013} \newline
    \citet{Sinoara:2014} \newline
    \citet{Peng:2015} \newline
    \citet{Wang:2015} \\ \hline

\textbf{Re8}          
  & Reuters-21578 & 7674    & 8901  & 8 
  &  \citet{Marcacini:2012} \newline
      \citet{Marcacini:2013}  \\ \hline

\textbf{Reuters-21578} 
  & Reuters-21578 & 21578 &  Not \newline Obtained  & 135 
  &  \citet{Cai:2014} \newline
     \citet{Sinoara:2014} \newline
     \citet{Peng:2015} \newline
     \citet{Wang:2015} \\ \hline

\textbf{Re0} 
  & Reuters-21578      & 1500   & 2886   & 13
  &  \citet{popat:17}   \\ \hline
  
\textbf{Re1} 
  & Reuters-21578      & 1650   & 3758  & 25
  & \citet{popat:17}   \\ \hline
  
\textbf{La1} 
  & TREC      & 3200  & 6188   & 6
  & \citet{popat:17}   \\ \hline

\textbf{La2} 
  & TREC      & 3040   & 6060   & 6
  & \citet{popat:17}   \\ \hline

\textbf{Classic} 
  & Medline, CACM      & 4089   & Not \newline Obtained   & 7
  &  \citet{popat:17}   \\ \hline

\end{tabularx}
\label{tab_datasets_trabalhos_relacionados}

\end{table*}
}

Reuters-21578 \citep{Dataset:Reuters21578} is one of the most widely used datasets. In some studies, the entire dataset has been used, whereas subsets of the database, such as Reu and Re8, have been used in other studies. This dataset consists of 21578 news articles from the Reuters Agency dating from 1987, which are classified into 135 categories. As can be observed in the table \ref{tab_datasets_trabalhos_relacionados}, the number of terms for this dataset could not be obtained from the studied papers.

Another commonly used dataset is the 20 newsgroups or 20ng \citep{20ng}; it contains 18808 e-mails sent through Internet discussion forums, and consists of 20 categories.

The Tdt dataset consists of 9824 news articles collected over six months, from January to June 1998, and classified into 193 categories. This information was obtained from six different sources.

Another commonly used document collection is the Hitech \citep{Karypis:2006}, which consists of news articles from the San Jose Mercury newspaper. The articles are classified into 6 categories: health, medicine, electronics, computers, technology, and research. This dataset contains 2301 documents.

In addition to these datasets, there are other ones, like Ohscal \citep{Dataset:Ohscal}, a subset of the Ohsumed dataset \citep{Hersh:1994} contains medical articles from 1987 to 1991, and dataset Afp, from the TREC-5 conference \citep{TREC5}, consists of a set of 695 Spanish articles categorized into 25 categories. More details about all these datasets can be found in their referenced works (Table \ref{tab_datasets_trabalhos_relacionados}).


\subsection{Clustering Evaluation Metrics}
\label{sec:metricas}

According to  \citet{Rijsbergen:1979}, the evaluation of the information retrieval systems (including clustering algorithms) is of great importance to users for social and economic issues. First, these evaluations help users to verify the need to utilize one of these systems by substituting the methods already in use and then selecting which one should be used. Further, the evaluation also helps users determine the cost of one of these systems and decide whether it is worth employing. Among the various features of an information retrieval system, Precision and Recall, and the metrics derived from them are highlighted when the aim is to demonstrate the effectiveness of the system.

The main metrics used in the studied papers are listed in Table \ref{sumario_metricas}. Some metrics use Precision and Recall values, which are then indirectly used by various studies. In Table \ref{sumario_metricas}, Precision and Recall are separated in order to highlight the studies that used the values from these metrics separately in their comparisons.

{
\renewcommand{\arraystretch}{2.0}
\linespread{1.0}
\begin{table*}[!ht]
  \centering
  \caption{Evaluation metrics used in the studies.}
  \begin{small}
  \setlength{\tabcolsep}{3pt}
  \scriptsize

  \begin{tabular}{p{6cm}p{0.3cm}p{5cm}}
  
  \textbf{Metrics} &  & \textbf{Studies where applied}  \\ \hline

  \textbf{Precision}  \newline
    \citep{Rijsbergen:1979, Marcacini:2012} 
    & & \citet{Huang:2011} 
    \newline
    \citet{zhiyi:2015}
    \newline
     \citet{protasiewicz:2016}  
     \newline
      \citet{hoxha:2016}
     \newline
    \citet{popat:17}
    \\ \hline

  \textbf{Recall} \newline
    \citep{Rijsbergen:1979, Marcacini:2012} 
    & &  \citet{Huang:2011} 
    \newline
     \citet{zhiyi:2015}
    \newline
     \citet{protasiewicz:2016}  
     \newline
     \citet{hoxha:2016}
     \newline
      \citet{popat:17}
    \\ \hline

  \textbf{Overall F-measure} \newline
    \citep{Rijsbergen:1979, Marcacini:2012, Larsen:1999, Zhao:2005}
    & & \citet{Garcia:2010a} \newline
     \citet{Garcia:2010b} \newline
      \citet{CorreaMorris:2010} \newline
       \citet{Huang:2011}\newline
     \citet{Marcacini:2012} \newline
       \citet{Marcacini:2013} \newline
      \citet{Sinoara:2014} \newline
       \citet{Peng:2015}
   	 \newline
   	  \citet{protasiewicz:2016}  
     \newline
      \citet{hoxha:2016}
     \newline
    \citet{popat:17}
     \\ \hline

  \textbf{FCubed} \newline
    \citep{Amigo:2009, Garcia:2010a}  
    & &  \citet{Garcia:2010a} \newline
      \citet{Garcia:2010b} \\ \hline

  \textbf{HF1} \newline
    \citep{Garcia:2010a}    
    & &  \citet{Garcia:2010a}   
    \newline
     \citet{irfan:16}
    \\ \hline

  \textbf{Overall F1-Travel} \newline
    \citep{Garcia:2010b}    
    & &  \citet{Garcia:2010b}   \\ \hline

  \textbf{Lexical F-measure} \newline
    \citep{irfan:16}        
    & & \citet{irfan:16} \\ \hline

  \textbf{ROUGE} \newline
    \citep{lin:03}      
    & &  \citet{Wang:2010}  \\ \hline

  \textbf{Accuracy} \newline
    \citep{Wang:2010, Wang:2015}     
    & & \citet{Wang:2010} \newline
    \citet{Wang:2015} \\ \hline

  \textbf{Relative Cost Ratio} \newline
    \citep{Cai:2014}        
    & &  \citet{Cai:2014} \\ \hline

  \end{tabular}
  \end{small}

  \label{sumario_metricas}
\end{table*}
}

In this study, a detailed description of only the most widely used metrics, or those with relevant characteristics for hierarchical clusters, is presented. The evaluation metrics Accuracy \citep{Wang:2010}, ROUGE \citep{lin:03}, and Relative Cost Ratio \citep{Cai:2014} are not detailed here, and additional information about them can be found in the corresponding works.

\subsubsection{Precision and Recall}

In the evaluation of clustering algorithms, the widely used metrics are those based on Precision and Recall values. In information retrieval systems, Precision and Recall are the relevance measures used to assess the retrieved data. Precision represents the fraction of the retrieved instances that are relevant, and Recall represents the fraction of the relevant instances that were retrieved. Thus, a high value of Precision implies that a large number of the returned instances are relevant, while a high value of Recall implies that a large amount of relevant data was returned \citep{Rijsbergen:1979}. In the context of clustering techniques, Precision and Recall are used in tests with previously classified datasets in order to compare this classification with the generated clusters. 

Let consider $C = \{C_1, \dots, C_n\}$ as the set of $n$ document clusters generated by a cluster algorithm, and $L = \{L_1, \dots, L_m\}$ as the set of $m$ real classes of the documents (previously assigned).

The Precision value (Equation \ref{eq_precision}) of cluster $C_i$ with respect to class $L_j$ represents the fraction of documents inserted in cluster $C_i$ that correspond to class $L_j$. The value can range from 0 to 1, and, if $P (C_i, L_j) = 1$, it implies that all the documents inserted into cluster $C_i$ correspond to class $L_j$. The Recall value (Equation \ref{eq_recall}) of $C_i$ with respect to $L_j$ represents the fraction of documents corresponding to class $L_j$ that are inserted in cluster $C_i$. This value also ranges from 0 to 1, and, if $R (C_i, L_j) = 1$, it implies that cluster $C_i$ contains all the documents from class $L_j$ \citep{Marcacini:2012}.

  \begin{equation}
    P (C_i, L_j) = \frac{ | C_i \cap L_j | } { | C_i | }
    \label{eq_precision}
  \end{equation}

  \begin{equation}
    R (C_i, L_j) = \frac{ | C_i \cap L_j | } { | L_j | }
    \label{eq_recall}
  \end{equation}

If $P (C_i, L_j)$ and $R (C_i, L_j)$ values are equal to 1, it implies that cluster $C_i$ corresponds exactly to class $L_j$. It can be observed that these values cannot independently represent the correspondence between cluster $C_i$ and class $L_j$, and therefore, various cluster evaluation metrics use a combination of these values. Some of these metrics are described in the following sections.

\subsubsection{F-Measure}
\label{sec:f_measure}

F-measure \citep{Rijsbergen:1979, Larsen:1999} (also known as F\textsubscript{SCORE} and F\textsubscript{1}) is a commonly used metric for the evaluation of clustering algorithms. The F-measure value for cluster $C_i$ with respect to class $L_j$ is defined by Equation \ref{eq_fmeasure}, which represents the harmonic average between Precision and Recall. The F-measure values can range from 0 to 1, and a larger value represents a greater correspondence between cluster $C_i$ and class $L_j$.

\begin{equation}
  F(C_i, L_j) = \frac{ 2 \times P (C_i, L_j) \times R (C_i, L_j) } 
            { P (C_i, L_j) + R (C_i, L_j) }
  \label{eq_fmeasure}
\end{equation}

The general $F$ value for class $L_j$ is obtained by using the highest value of $F (C_i, L_j)$ (Equation \ref{eq_fmeasure_grupo}) and considering all the clusters $C_i \in C$, i.e., the value of $F(L_j)$ is calculated based on the cluster having the greatest correspondence with class $L_j$.

Thus, the general F-measure value for the clustering problem, including all the clusters $C_1, \dots, C_n$, generated from a set of documents, is given by Equation \ref{eq_fmeasure_geral} \citep{Marcacini:2012, Zhao:2005}. It is the average of the $F(L_j)$ values weighted by the number of documents found in the corresponding classes. Various authors refer to this general measure as Overall F\textsubscript{1} \citep{Garcia:2010a, Garcia:2010b} , Overall F-measure \citep{CorreaMorris:2010, Larsen:1999}, F\textsubscript{SCORE} \citep{Marcacini:2012, Marcacini:2013, Sinoara:2014}, and F-measure \citep{Huang:2011, Peng:2015}.

\begin{equation}
  F (L_j) = max_{\substack{ \\ C_i \in C }} F(C_i, L_j)
  \label{eq_fmeasure_grupo}
\end{equation} 

\begin{equation}
  Overall\, \textit{F-measure} = \frac{  \sum_{j=1}^{m} (|L_j| F(L_j)) } { \sum_{j=1}^{m} |L_j| } 
  \label{eq_fmeasure_geral}
\end{equation}

It can be observed that during the calculation of F-measure in the assessment of hierarchical clustering algorithms, that create clusters and sub-clusters, for calculating the values of Recall and Precision of one cluster, the documents in its sub-cluster are also considered \citep{Larsen:1999}.

\subsubsection{FCubed}

The FCubed \citep{Garcia:2010a} metric is suitable for situations that may involve intersections between the generated clusters. If two documents correspond to two classes simultaneously, in an ideal clustering, it would be expected that they would be found together in two clusters. Thus, this metric aims to evaluate how closely the generated clusters meet this expectation. This metric is an adaptation of the F-measure metric using Precision BCubed and Recall BCubed \citep{Amigo:2009}. First, the Multiplicity Precision (Equation \ref{eq_multiplicity_precision}) and Multiplicity Recall (Equation \ref{eq_multiplicity_recall}) values are calculated. In the following equations, let us consider that for two documents $d_i$ and $d_j$, $L_{ij}$ and $C_{ij}$ are the set of classes and the set of clusters, respectively, that contain both documents. For example, assuming that there are documents $d_1$, $d_2$, $d_3$ and $d_4$, classes $l_1 = \{d_1, d_2, d_4\}$, $l_2 = \{d_1, d_3\}$, $l_3 = \{d_1, d_2, d_3\}$ and $l_4 = \{d_2, d_4\}$, and  clusters $c_1 = \{d_1, d_2, d_3\}$, $c_2 = \{c_2, c_4\}$ and $c_3 = \{c_3, c_4\}$, then $L_{12} = \{l_1, l_3\}$ and $C_{12} = \{c_1\}$.

  \begin{equation}
    MP (d_i, d_j) = \frac{ min ( | C_{ij} | , | L_{ij} | ) } { | C_{ij} | }
    \label{eq_multiplicity_precision}
  \end{equation}

  \begin{equation}
    MR (d_i, d_j) = \frac{ min ( | C_{ij} | , | L_{ij} | ) } { | L_{ij} | }
    \label{eq_multiplicity_recall}
  \end{equation}

Thus, the Precision BCubed value (Equation \ref{eq_precision_bcubed}) is given by the mean of $MP(d_i, d_j)$ for all the documents $d_i$ and $d_j$, and Recall BCubed (Equation \ref{eq_recall_bcubed}) is calculated in a similar manner by using $MR(d_i, d_j)$. The value of FCubed is given by Equation \ref{eq_fcubed}.
ca
\begin{equation}
  Precision\ BCubed (d_i, d_j) = Avg_{d_i} \Big[ Avg_{d_j . C_{ij} \neq \emptyset } [MP (d_i, d_j)] \Big]
  \label{eq_precision_bcubed}
\end{equation}

\begin{equation}
  Recall\ BCubed (d_i, d_j) = Avg_{d_i} \Big[ Avg_{d_j . L_{ij} \neq \emptyset } [MR (d_i, d_j)] \Big]
  \label{eq_recall_bcubed}
\end{equation}

\begin{equation}
  FCubed = \frac{ 2 \times Precision\ BCubed \times Recall\ BCubed } 
          { Precision\ BCubed + Recall\ BCubed }
  \label{eq_fcubed}
\end{equation}

It can be observed that an increase in the value of \textit{FCubed} results in an increase in the ability of the algorithm to capture the relationship between the documents that share the same classes by inserting these documents into the same cluster.

\subsubsection{HF1}

HF1, a metric proposed by \citet{Garcia:2010a}, is based on the F-measure and is suitable for hierarchical clustering because it includes the generated hierarchy in its calculations.

Considering the hierarchy composed of $N$ manually classified topics, the structure could be represented according to the ancestors of each topic, the form of structure of the ancestors can be formally showed by $TH = \{(t_1, a_1^1), \dots, (t_1, a_{k_1}^1), \dots, (t_N, a_1^N), \dots, (t_N, a_{k_N}^N)\}$, $A(t_i) = \{ a_1^i, \dots, a_{k_i}^i \}$ is the set of $k_i$ ancestors for the topic $t_i$. Similarly, the hierarchy of clusters generated by an algorithm can be represented by the following: $CH=\{(c_1, a_1^1)$ $, \dots, (c_1, a_{m_1}^1),$ $ \dots, (c_M, a_1^M),$ $ \dots, (c_M, a_{m_M}^M)\}$, where $c_i$ is one of the $M$ clusters generated and $A(c_i) = \{ a_1^i, \dots, a_{m_i}^i \}$ is the set of their ancestors in the hierarchy.

Then, the value of $\sigma(t_i)$ (Equation \ref{eq_sigma}) can be defined. It is a cluster $c_i \in C$ that returns the greatest value of $F(c_i, t_i)$. It must be noted that $\sigma(c_i)$ is a cluster and not a numerical value.

\begin{equation}
  \sigma(t_i)=arg\ max_{c_i \in C}{ F(c_i, t_i) }
  \label{eq_sigma}
\end{equation}

It is also important to define the value of $CP$ that is used in the Equations \ref{eq_hprecision} and \ref{eq_hrecall}; it is the number of pairs in the topics  hierarchy $TH$ that are correctly identified in the clusters hierarchy $CH$. Given a pair of topics $(t_i, t_j) \in TH$, this pair is considered to be correctly identified if there is a pair of clusters $(\sigma(t_i), \sigma(t_j)) \in CH$. Thus, if two topics have an ancestral relationship, their corresponding clusters should also exhibit this relationship.

Thus, the value of $HF1$ (Equation \ref{eq_hf1}) is defined as a function of $HPrecision$ (Equation \ref{eq_hprecision}) and $HRecall$ (Equation \ref{eq_hrecall}).

\begin{equation}
  HPrecision = \frac { CP } { |CH| }
  \label{eq_hprecision}
\end{equation}

\begin{equation}
  HRecall = \frac { CP } { |TH| }
  \label{eq_hrecall}
\end{equation}

\begin{equation}
  HF1= \frac{ 2 \times HPrecision \times HRecall } {HPrecision + HRecall}
  \label{eq_hf1}
\end{equation}

This metric differs from the others because it is specific to hierarchical clustering and aims to assess the ancestral relationship between the clusters in the hierarchy. It can be observed that, to use this metric, the previously classified data used for this test must be organized in a hierarchical form (hierarchy of topics).

\subsubsection{Overall F1-Travel}

The Overall F1-Travel metric presented by \citet{Garcia:2010b} is also suitable for the evaluation of hierarchical clustering. This metric is fairly similar to the Overall F-Measure; however, it aims to evaluate the navigability of the generated hierarchy. Extremely vertical and extremely horizontal hierarchies can hinder the search for a document, and this metric aims to evaluate the cost of navigation to find the topics. Equation \ref{eq_ftravel} shows the calculation of $\textit{F1-Travel}(L_j)$, which represents the cost of navigation to find the topic $L_j$. In this equation, $n$ represents the total number of documents, and $v$ denotes the number of clusters visited in the hierarchy during the search from the root for cluster $\sigma(L_j)$. The cluster $\sigma(L_j)$ is the ``best matching'' cluster with the topic $L_j$; in other words, $\sigma(L_j)$ is the cluster $C_i$ that maximizes $F(C_i, L_j)$ (equation \ref{eq_fmeasure}). 
In order to determine $v$, a best-first search is used, as detailed by \citet{Garcia:2010b}. The Overall \textit{F1-Travel} value can be defined by Equation \ref{eq_overall_ftravel}.

\begin{equation}
  \textit{F1-Travel} (L_j) = F(L_j, \sigma(L_j))(1 - \frac{v}{2n})
  \label{eq_ftravel}
\end{equation}

\begin{equation}
  Overall\, \textit{F1-Travel} = \frac{ \sum_{j=1}^{m} (|L_j| \textit{F1-Travel}(L_j)) } 
                    { \sum_{j=1}^{m} |L_j| } 
  \label{eq_overall_ftravel}
\end{equation}


\subsection{Baseline Algorithms}
\label{sec:algoritmos_comparados}

Another point of difference among the experiments conducted in the studies is the algorithms used as baseline for the comparison and evaluation of the proposed methods. Table \ref{tab_algoritmos_comparados} shows the baseline algorithms used.

{
\renewcommand{\arraystretch}{2.0}
\linespread{0.9}
\begin{table*}[!ht]
\centering
\caption{Baseline algorithms.}
  \begin{small}
  \setlength{\tabcolsep}{3pt}
  \scriptsize
  \resizebox {0.55\textwidth }{!}{ %
\begin{tabular}{p{6cm}p{5cm}}

\textbf{Algorithm / Strategy}         & \textbf{Studies where applied}        \\ \hline 

\textbf{UPGMA} \newline \citep{Jain:1988} 
  & \citet{Garcia:2010a} \newline
    \citet{Garcia:2010b}     \\ \hline

\textbf{Bisecting K-means} \newline \citep{Karypis:2000, Zhao:2005} 
  & \citet{Garcia:2010a} \newline
    \citet{Garcia:2010b} \newline
     \citet{Marcacini:2012} \newline
    \citet{Sinoara:2014}   \\ \hline

\textbf{Single-link} \newline \citep{Jain:1988, Sibson:1973} 
  & \citet{CorreaMorris:2010} \newline
     \citet{Peng:2015}    \\ \hline

\textbf{Complete-link} \newline \citep{Jain:1988, Defays:1977} 
  &  \citet{CorreaMorris:2010} \newline
     \citet{Peng:2015}    \\ \hline

\textbf{Star} \newline \citep{Aslam:2004} 
  & \citet{CorreaMorris:2010}            \\ \hline

\textbf{Extended Star} \newline \citep{Garcia:2003} 
  &  \citet{CorreaMorris:2010}            \\ \hline

\textbf{ACONS} \newline \citep{Alonso:2007} 
  & \citet{CorreaMorris:2010}            \\ \hline

\textbf{Incremental clustering using Indexing Trees} \newline \citep{Zhang:2007}
  &  \citet{LWang:2010}                 \\ \hline

\textbf{Title words based clustering} \newline \citep{Dai:2010} 
  & \citet{Huang:2011}                 \\ \hline

\textbf{Buckshot} \newline \citep{Cutting:1992} 
  & \citet{Marcacini:2012} \newline
    \citet{Marcacini:2013}   \\ \hline

\textbf{Average Random Clustering} \newline \citep{Marcacini:2012}
  & \citet{Marcacini:2012}             \\ \hline

\textbf{GFIO (K-Medians e K-means)} \newline \citep{Lin:2006} 
  & \citet{Cai:2014}                 \\ \hline

\textbf{Small-Space} \newline \citep{Guha:2003} 
  & \citet{Cai:2014}                 \\ \hline

\textbf{CLIQUE} \newline \citep{Agrawal:1998} 
  &  \citet{Peng:2015}                  \\ \hline

\textbf{K-means} \newline \citep{Macqueen:1967} 
  & \citet{Peng:2015} \newline
   \citet{kobren:17} \newline
   \citet{sangaiah:18} \newline
   \citet{popat:17} \\ \hline

\textbf{Support Vector Machine (SVM)} \newline \citep{Cortes:1995} 
  &  \citet{Wang:2015}                  \\ \hline

\textbf{Naive Bayes (NB)} \newline \citep{Hassan:2011} 
  & \citet{Wang:2015}                  \\ \hline
  
\textbf{Taxonomy Generation Process (TGP)} \newline \citep{camina:10} 
  & \citet{irfan:16}                  \\ \hline
  
\textbf{Hierarchical Agglomerative Clustering (HAC)} \newline \citep{hastie:05} 
  & \citet{protasiewicz:2016}
  \newline
   \citet{hoxha:2016} \newline
  \citet{kobren:17}
  \\ \hline
 
\textbf{BIRCH} \newline \citep{Zhang:96} 
  & 
   \citet{kobren:17}
  \\ \hline
 
\textbf{Mini-batch HAC (MB-HAC)} \newline \citep{kobren:17} 
  & 
  \citet{kobren:17}
  \\ \hline
 
\textbf{Stream K-means++ (SKM++)} \newline \citep{ackermann:12} 
  & 
  \citet{kobren:17}
  \\ \hline
 
\textbf{Mini-batch K-means (MB-KM)} \newline \citep{sculley:10} 
  & 
  \citet{kobren:17}
  \\ \hline
 
\textbf{BICO} \newline \citep{fichtenberger:13} 
  & 
   \citet{kobren:17}
  \\ \hline

\textbf{DBSCAN} \newline \citep{ester:96} 
  & 
  \citet{kobren:17}
  \\ \hline

\textbf{Hierarchical K-means(HKMeans)} \newline \citep{kobren:17} 
  & 
   \citet{kobren:17}
  \\ \hline

\textbf{Partitioning Around Medoids
(PAM)} \newline \citep{kaufman:09} 
  & 
  \citet{amelin:18}
  \\ \hline

\textbf{STREAM} \newline \citep{guha:03} 
  & 
  \citet{khalilian:16}
  \\ \hline

\textbf{Evolutionary Hierarchical Clustering Algorithm} \newline \citep{chakrabarti:06} 
  & 
  \citet{wang:13}
  \\ \hline

\end{tabular}}
\end{small}
\label{tab_algoritmos_comparados}

\end{table*}
}

These algorithms use to be algorithms that are common on the state of art to the problem it studies to compare its quality. 
It can be observed that the presented studies are significantly different in this respect, thus hindering a direct comparison between them. Some of the algorithms in Table \ref{tab_algoritmos_comparados} are not incremental or hierarchical, such as the K-means algorithm. Metrics such as the Overall F-measure can be applied to partitional or hierarchical algorithms; however, other metrics (such as HF1 and F1-travel) cannot be used for partitional algorithms, thus limiting the possibilities for a direct comparison.


\subsection{Discussion}
\label{sec:discussao}

An analysis of the characteristics of the studied papers reveals the differences with respect to various aspects. The studies that involve the application of a technique present a variety of final objectives for the use of clustering: the summarization of documents, organization of documents, identification of the relationship between topics, and identification of hot topics.

The studies that proposed new algorithms or techniques used various strategies with different focus areas. In general, the studies aimed to generate good-quality hierarchies of clusters; however, some studies proposed the generation of hierarchies that are easy to navigate, creation of meaningful levels in the hierarchy, or clustering that is not sensitive to the order in which the data is read. The adopted strategies focused on the selection of features, the use of different criteria for the hierarchy levels, or the use of privileged information and named entities.

With regard to the similarity and distance measures used, as discussed in the Section \ref{sec:medidas_similaridade}, the choice of the measure to be used may not be straightforward and depends on the context. However, Cosine Similarity is predominantly used because its properties are favorable to text clustering.

The datasets used in the studies were also significantly different. The most widely used databases are Reuters-21578, and its subsets Reu and Re8. This dataset has a large volume of data, which may help in the development and testing of more robust algorithms.

Various clustering evaluation metrics are used in the studies; Overall F-measure is used in a majority of studies. This metric can be used for hierarchical or partitional clustering algorithms, thus facilitating the comparison between the algorithms. The metrics HF1 and Overall F1-Travel are also highlighted. These metrics were presented by two of the studied works, and focus on the evaluation hierarchies.

Based on the algorithms used for comparison in the studies, it can be noted that none of these studies performed comparisons with each other; instead, the comparisons were made with studies conducted prior to the period encompassing the previous five years.


\section{Conclusions}
\label{sec:conclusoes}

This study collected published studies between the years 2010 and 2018 that were related to incremental and hierarchical text clustering. In order to select these works, a systematic review was conducted, obtaining some studies that apply these techniques in different scenarios and others that aim to develop new techniques and approaches. A survey of the literature available in this field has been given to provide a comprehensive insight into these topics.

An analysis of the characteristics of the studied papers reveals some  differences related to various aspects. The studies that involve the application of a technique present a variety of final objectives for the use of clustering: the  text summarization, organization of documents, identification of the relationship between topics, and identification of hot topics. The studies that proposed new algorithms or techniques used various strategies with different focus areas. In general, the studies aimed to generate good-quality hierarchies of clusters; however, some studies proposed the generation of hierarchies that are easy to navigate, creation of meaningful levels in the hierarchy, or clustering that is not sensitive to the order in which the data is read. The adopted strategies focused on the selection of features, the use of different criteria for the hierarchy levels, or the use of privileged information and named Entities. 

Based on the works that presented new clustering techniques, new approaches are being developed to outperform in time, accuracy and memory limitations. Those studies focused on different areas and adopted different strategies, therefore we can observe that the problem of incremental hierarchical text clustering has not been completely addressed, and there is no definitive strategy for this challenge.

As future work, comparative tests can be conducted for the presented algorithms to determine the algorithms that generate better hierarchies of clusters in dynamic environments. Earlier studies have not performed such comparisons. Such comparative study can also enable the development of new hybrid algorithms by combining the different characteristics of existing algorithms.

Incremental and hierarchical document clustering algorithms have been the focus of recent research, and it can be observed that they have great relevance and applicability to the current scenario. This study demonstrated that this field requires further exploration, various challenges must be overcome and a consensus on the best techniques to be used must be investigated.

\section*{Acknowledgements}
\label{Ack}
  We would like to thank CNPq,  FAPEMIG and CAPES (Brazilian agencies) for partial financial support.

\section*{Reference}

\bibliographystyle{elsarticle-num-names} 
\bibliography{refs}





\end{document}